\documentclass[sigconf]{acmart}
\AtBeginDocument{%
  \providecommand\BibTeX{{%
    \normalfont B\kern-0.5em{\scshape i\kern-0.25em b}\kern-0.8em\TeX}}}

\copyrightyear{2024}
\acmYear{2024}
\setcopyright{acmlicensed}\acmConference[WWW '24 Companion]{Companion Proceedings of the ACM Web Conference 2024}{May 13--17, 2024}{Singapore, Singapore}
\acmBooktitle{Companion Proceedings of the ACM Web Conference 2024 (WWW '24 Companion), May 13--17, 2024, Singapore, Singapore}
\acmDOI{10.1145/3589335.3651955}
\acmISBN{979-8-4007-0172-6/24/05}

\usepackage{multirow}
\usepackage{graphicx}

\begin{document}

\title[How Reliable is Your Simulator? Analysis on the Limitations of Current LLM-based User Simulators]{How Reliable is Your Simulator? Analysis on the Limitations of Current LLM-based User Simulators for Conversational Recommendation}


\author{Lixi Zhu}
\affiliation{%
  \institution{School of Computer Science and Technology, Beijing Jiaotong University}
  \city{Beijing}
  \country{China}
  \postcode{100044}}
\email{zlxxlz1026@gmail.com}

\author{Xiaowen Huang}
\authornote{corresponding author}
\affiliation{%
  \institution{School of Computer Science and Technology, Beijing Jiaotong University}
  \institution{Beijing Key Lab of Traffic Data Analysis and Mining
, Beijing Jiaotong University}
  \institution{Key Laboratory of Big Data \& Artificial Intelligence in Transportation(Beijing Jiaotong University), Ministry of Education}
  \city{Beijing}
  \country{China}
  \postcode{100044}}
\email{xwhuang@bjtu.edu.cn}

\author{Jitao Sang}
\affiliation{%
  \institution{School of Computer Science and Technology, Beijing Jiaotong University}
  \institution{Beijing Key Lab of Traffic Data Analysis and Mining
, Beijing Jiaotong University}
  \institution{Key Laboratory of Big Data \& Artificial Intelligence in Transportation(Beijing Jiaotong University), Ministry of Education}
  \city{Beijing}
  \country{China}
  \postcode{100044}}
\email{jtsang@bjtu.edu.cn}

\renewcommand{\shortauthors}{Lixi Zhu, Xiaowen Huang, \& Jitao Sang}

\begin{abstract}
Conversational Recommender System (CRS) interacts with users through natural language to understand their preferences and provide personalized recommendations in real-time. CRS has demonstrated significant potential, prompting researchers to address the development of more realistic and reliable user simulators as a key focus. Recently, the capabilities of Large Language Models (LLMs) have attracted a lot of attention in various fields. Simultaneously, efforts are underway to construct user simulators based on LLMs. While these works showcase innovation, they also come with certain limitations that require attention. In this work, we aim to analyze the limitations of using LLMs in constructing user simulators for CRS, to guide future research. To achieve this goal, we conduct analytical validation on the notable work, iEvaLM. Through multiple experiments on two widely-used datasets in the field of conversational recommendation, we highlight several issues with the current evaluation methods for user simulators based on LLMs: (1) Data leakage, which occurs in conversational history and the user simulator's replies, results in inflated evaluation results. (2) The success of CRS recommendations depends more on the availability and quality of conversational history than on the responses from user simulators. (3) Controlling the output of the user simulator through a single prompt template proves challenging. To overcome these limitations, we propose SimpleUserSim, employing a straightforward strategy to guide the topic toward the target items. Our study validates the ability of CRS models to utilize the interaction information, significantly improving the recommendation results. 
\end{abstract}



\maketitle
\section{Introduction}
Recommender systems have played a pivotal role in enhancing performance and customer satisfaction across diverse industries. However, traditional recommender systems, which learn from offline data, can only capture users' long-term preferences through their historical behaviors \cite{he2017neural,koren2009matrix}. These traditional systems fail to address two important questions: what is the user's current preference, and which specific aspects of an item does the user favor? \cite{gao2021advances}. Conversational Recommender System (CRS) \cite{sun2018conversational}, typically comprising a dialogue module and a recommendation module, can capture not only the user's long-term preferences but also their real-time preferences. This is achieved through multiple rounds of natural language interactions with the user, enabling the system to make personalized recommendations in real-time. Despite the considerable success of current CRS, the challenge of constructing a realistic and trustworthy user simulator for evaluating the CRS's performance remains unaddressed \cite{friedman2023leveraging}. Current CRS can be divided into two categories: attribute-based CRS \cite{lei2020estimation, lei2020interactive, deng2021unified, zhao2022knowledge} and natural language processing (NLP)-based CRS \cite{chen2019towards, zhou2020improving, wang2022barcor, wang2022towards}. The attribute-based CRS is designed with a policy module that controls the actions of the CRS in each round of interaction, aiming to achieve successful recommendations by understanding users' real-time preferences with the fewest possible interactions. On the other hand, NLP-based CRS prioritizes offering users a seamless conversation experience. They also incorporate information related to the recommended items into the response text, enhancing the interpretability of the recommendation results. In attribute-based CRS, the user simulator's responses are based on fixed templates, neglecting the flow of the conversation. In contrast, NLP-based CRS takes into account the flow of the conversation, but evaluations are based on fixed conversations, which may neglect the interactivity of conversational recommendations. 

Recently, new opportunities have arisen from the development of the Large Language Models (LLMs) \cite{zhao2023survey}. LLMs can utilize world knowledge and commonsense reasoning for text comprehension and generation, in some aspects approaching human-level intelligence \cite{saparov2023testing, xi2023rise}. This has prompted researchers to explore the potential of employing LLMs in various tasks within the recommendation domain \cite{dai2023uncovering, gao2023chat, bao2023tallrec, hou2023large, li2023large, he2023large}, particularly in the construction of user simulators, which has attracted significant attention \cite{wang2023rethinking, wang2023recagent, zhang2023generative}. Recent studies have utilized LLMs as user simulators to evaluate the performance of CRS \cite{wang2023rethinking, huang2023recommender}, overcoming the template-based limitations of previous research and demonstrating promising results. However, the majority of current studies in constructing user simulators offer session-level guidance to LLMs on generating responses via a single prompt template, which we refer to as `single-prompt’. Although these studies showcase innovation, they also present certain limitations that need to be addressed.

The primary differences in user simulators constructed by various studies lie in the granularity of the prompts used. In our work, we conduct analytical validation of the notable study iEvaLM \cite{wang2023rethinking}. 
\newline
Through multiple experiments on two widely-used datasets in the field of conversational recommendation, we identified the following major findings after empirical experiments:
\begin{itemize}
    \item Data leakage, which occurs in conversational history and the user simulator's replies, results in inflated evaluation results.
    \item The success of CRS recommendations depends more on the availability and quality of conversational history than on the responses from user simulators.
    \item Controlling the output of the user simulator through a single prompt template proves challenging.
\end{itemize}

These findings reveal the limitations inherent in using LLMs for constructing user simulators for CRS. Moreover, we introduce SimpleUserSim, which employs a straightforward strategy to guide the topic toward the target items. Our research validates the CRS model's capability to leverage interaction information, thereby significantly enhancing the quality of recommendation results.

\section{LLM as USER SIMULATOR for CRS}
\subsection{Workflow}
\setlength{\belowcaptionskip}{-5pt}
\begin{figure}[htp]
    \includegraphics[width=0.5\textwidth]{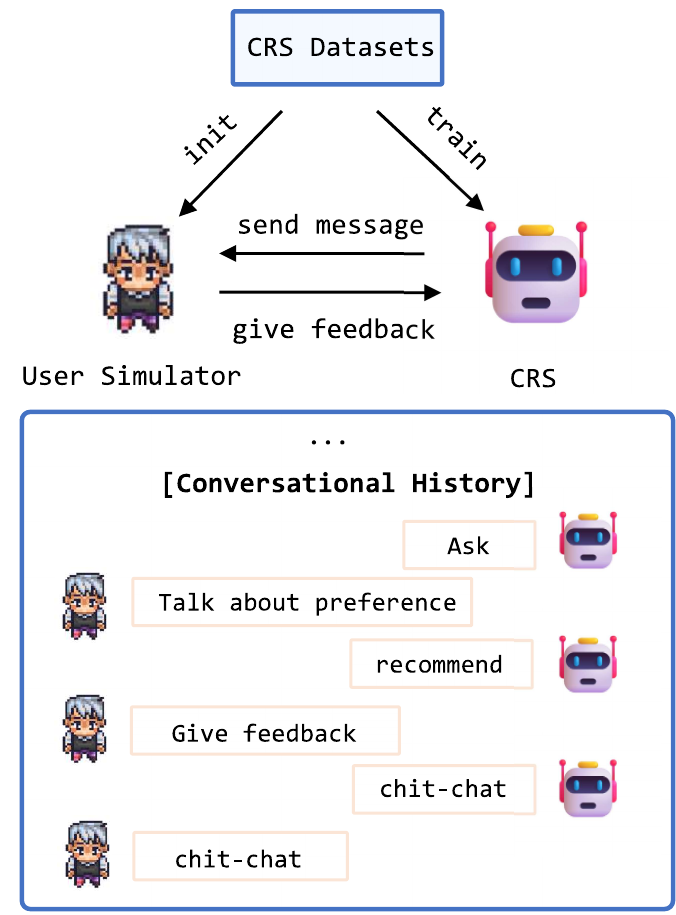}
    \caption{Workflow of the User Simulator.}
    \label{fig-1}
\end{figure}
The workflow of the user simulator is illustrated in Figure \ref{fig-1}. The user simulator is initialized based on existing CRS datasets, which contain human-annotated conversations and target item titles. The user simulator takes the target items as real-time user preferences and uses natural language to interact with the CRS based on human-annotated conversational history, expressing real-time preferences to obtain the desired recommendations.

\subsection{Experimental setup}
Our work builds upon iEvaLM. For detailed information about our experimental settings, please refer to iEvaLM\footnote{https://github.com/RUCAIBox/iEvaLM-CRS/}.
\newline
\textbf{Dataset:} We conduct experiments on two classic datasets in the conversational recommendation domain: ReDial \cite{li2018towards} and OpenDialKG \cite{moon2019opendialkg}. ReDial is a movie conversational recommendation dataset that contains 10,006 conversations. OpenDialKG is a multi-domain conversational recommendation dataset with 13,802 conversations, and only the movie domain was utilized in our experiments.
\newline
\textbf{Baselines:} We conduct comparative experiments using three classical approaches in the field of conversational recommendations, as well as with a well-known large language model named ChatGPT:
\begin{itemize}
    \item KBRD \cite{chen2019towards}: It utilizes an external Knowledge Graph (KG) to enhance the semantics of entities mentioned in the conversational history.
    \item BARCOR \cite{wang2022barcor}: It proposes a unified CRS based on BART \cite{lewis2019bart}, which tackles two tasks using a single model.
    \item UniCRS \cite{wang2022towards}: It proposes a unified CRS model that leverages knowledge-enhanced prompt learning.
    \item ChatGPT \footnote{https://chat.openai.com/}: We employ the publicly available model GPT-3.5-turbo-0613 provided by the OpenAI API.
\end{itemize}
\textbf{Evaluation Metrics:} Following existing work, we adopt Recall@$k$ to evaluate the recommendation task. Similarly, we set $k = 1, 10, 50$, for both the ReDial and OpenDialKG datasets. Additionally, following existing work, we set the maximum number of interaction turns between the user simulator and the CRS at $t = 5$.

\subsection{Experiment}
During our experiments, we observed the `data leakage', where the titles of the target items were unintentionally disclosed. Consequently, we aim to ascertain how many successful recommendation conversations are affected by data leakage. Additionally, considering the influence of conversational history in datasets, we aim to quantify the contribution of the user simulator to the success of recommendation conversations. It is noteworthy that iEvaLM primarily focuses on expressing preferences when the CRS inquires about user preferences or provides feedback upon making a recommendation. We aim to determine whether its output aligns with expectations. Based on these considerations, we conduct experiments to evaluate iEvaLM, aiming to address the following research questions:
\begin{itemize}
    \item \textbf{RQ1:} Does the current user simulator, iEvaLM, exhibit data leakage, and if so, in which process? How does the model perform when we ignore these successful recommendations affected by data leakage?
    \item \textbf{RQ2:} How much do successful recommendation conversations depend on user simulator interactions compared to conversational history?
    \item \textbf{RQ3:} Can the user simulator generate responses that meet expectations across various dataset scenarios? If not, why?
\end{itemize}

\begin{figure}[h]
    \includegraphics[width=0.47\textwidth]{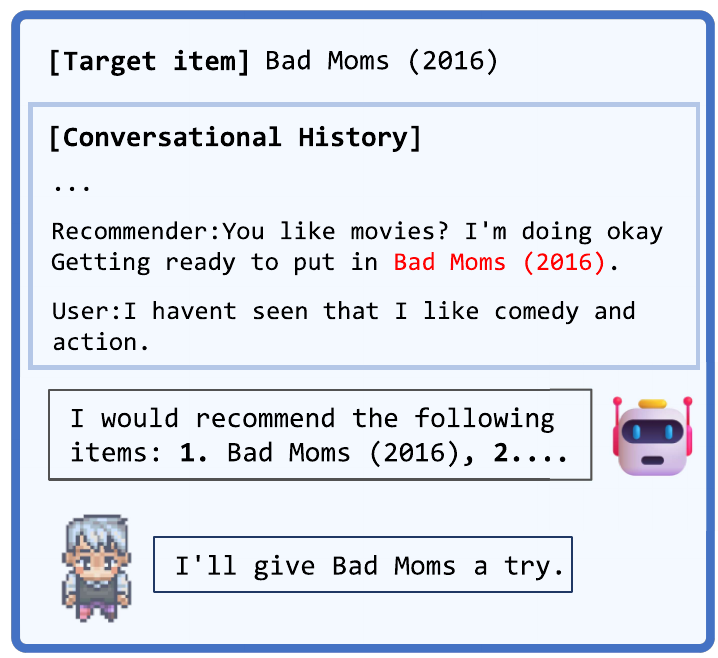}
    \caption{Data leakage from conversational history leads to successful recommendations.}
    \label{fig-2}
\end{figure}
\begin{figure}[h]
    \includegraphics[width=0.47\textwidth]{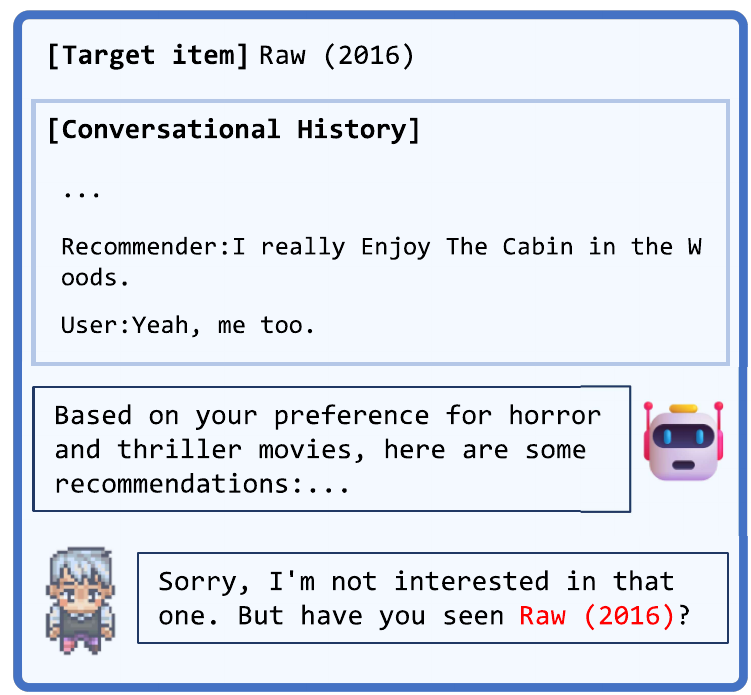}
    \caption{Data leakage from user simulator leads to successful recommendations.}
    \label{fig-3}
\end{figure}
In RQ1, when we ignore these successful recommendation conversations affected by data leakage, model performance deteriorates further, indicating that the impact of data leakage is significant. This also suggests that the evaluation of the original user simulator is not reasonable, and we should avoid such leakage to construct a more realistic and trustworthy user simulator. In RQ2, if the CRS successfully recommends on the first turn, it implies that the CRS can make successful recommendations using only information from conversational history. However, if it is not successful, it indicates that the CRS can effectively utilize information from interactions with the user simulator. In RQ3, a very low percentage of successful recommendations across multiple rounds of interaction may suggest that the user simulator does not generate responses that meet our expectations.

\begin{table*}[htb]
\caption{Performance of existing CRS methods and ChatGPT under iEvaLM in various data leakage scenarios.}
\resizebox{\textwidth}{!}{
\begin{tabular}{cc|ccc|ccc|ccc|ccc}
\hline
\multicolumn{2}{c|}{Model}                                                                & \multicolumn{3}{c|}{KBRD}                                                                                                                                                         & \multicolumn{3}{c|}{BARCOR}                                                                                                                                                       & \multicolumn{3}{c|}{UniCRS}                                                                                                                                                       & \multicolumn{3}{c}{ChatGPT}                                                                                                                                                       \\ \hline
\multicolumn{2}{c|}{datasets}                                                             & Recall@1                                                  & Recall@10                                                 & Recall@50                                                 & Recall@1                                                  & Recall@10                                                 & Recall@50                                                 & Recall@1                                                  & Recall@10                                                 & Recall@50                                                 & Recall@1                                                  & Recall@10                                                 & Recall@50                                                 \\ \hline
\multirow{4}{*}{ReDial}     & iEvaLM                                                      & 0.033                                                     & 0.229                                                     & 0.575                                                     & 0.032                                                     & 0.190                                                     & 0.499                                                     & 0.219                                                     & 0.454                                                     & 0.718                                                     & 0.220                                                     & 0.539                                                     & 0.816                                                     \\
                            & \begin{tabular}[c]{@{}c@{}}iEvaLM\\ (-history)\end{tabular} & \begin{tabular}[c]{@{}c@{}}0.016\\ (-51.5\%)\end{tabular} & \begin{tabular}[c]{@{}c@{}}0.187\\ (-18.3\%)\end{tabular} & \begin{tabular}[c]{@{}c@{}}0.542\\ (-5.74\%)\end{tabular} & \begin{tabular}[c]{@{}c@{}}0.032\\ (-0.0\%)\end{tabular}  & \begin{tabular}[c]{@{}c@{}}0.178\\ (-6.3\%)\end{tabular}  & \begin{tabular}[c]{@{}c@{}}0.473\\ (-5.2\%)\end{tabular}  & \begin{tabular}[c]{@{}c@{}}0.204\\ (-6.8\%)\end{tabular}  & \begin{tabular}[c]{@{}c@{}}0.429\\ (-5.5\%)\end{tabular}  & \begin{tabular}[c]{@{}c@{}}0.587\\ (-18.2\%)\end{tabular} & \begin{tabular}[c]{@{}c@{}}0.198\\ (-10.0\%)\end{tabular} & \begin{tabular}[c]{@{}c@{}}0.524\\ (-2.8\%)\end{tabular}  & \begin{tabular}[c]{@{}c@{}}0.833\\ (+2.1\%)\end{tabular}  \\
                            & \begin{tabular}[c]{@{}c@{}}iEvaLM\\ (-response)\end{tabular}  & \begin{tabular}[c]{@{}c@{}}0.033\\ (-0.0\%)\end{tabular}  & \begin{tabular}[c]{@{}c@{}}0.203\\ (-11.4\%)\end{tabular} & \begin{tabular}[c]{@{}c@{}}0.506\\ (-12.0\%)\end{tabular} & \begin{tabular}[c]{@{}c@{}}0.032\\ (-0.0\%)\end{tabular}  & \begin{tabular}[c]{@{}c@{}}0.201\\ (+5.8\%)\end{tabular}  & \begin{tabular}[c]{@{}c@{}}0.465\\ (-6.8\%)\end{tabular}  & \begin{tabular}[c]{@{}c@{}}0.081\\ (-63.0\%)\end{tabular} & \begin{tabular}[c]{@{}c@{}}0.311\\ (-31.5\%)\end{tabular} & \begin{tabular}[c]{@{}c@{}}0.644\\ (-10.3\%)\end{tabular} & \begin{tabular}[c]{@{}c@{}}0.149\\ (-32.3\%)\end{tabular} & \begin{tabular}[c]{@{}c@{}}0.387\\ (-28.2\%)\end{tabular} & \begin{tabular}[c]{@{}c@{}}0.670\\ (-17.9\%)\end{tabular} \\
                            & \textbf{\begin{tabular}[c]{@{}c@{}}iEvaLM\\ (-both)\end{tabular}}    & \textbf{\begin{tabular}[c]{@{}c@{}}0.011\\ (-66.7\%)\end{tabular}} & \textbf{\begin{tabular}[c]{@{}c@{}}0.143\\ (-37.6\%)\end{tabular}} & \textbf{\begin{tabular}[c]{@{}c@{}}0.451\\ (-21.6\%)\end{tabular}} & \textbf{\begin{tabular}[c]{@{}c@{}}0.032\\ (-0.0\%)\end{tabular}}  & \textbf{\begin{tabular}[c]{@{}c@{}}0.187\\ (-1.6\%)\end{tabular}}  & \textbf{\begin{tabular}[c]{@{}c@{}}0.430\\ (-13.8\%)\end{tabular}} & \textbf{\begin{tabular}[c]{@{}c@{}}0.029\\ (-86.8\%)\end{tabular}} & \textbf{\begin{tabular}[c]{@{}c@{}}0.245\\ (-46.0\%)\end{tabular}} & \textbf{\begin{tabular}[c]{@{}c@{}}0.621\\ (-13.5\%)\end{tabular}} & \textbf{\begin{tabular}[c]{@{}c@{}}0.044\\ (-80.0\%)\end{tabular}} & \textbf{\begin{tabular}[c]{@{}c@{}}0.271\\ (-49.7\%)\end{tabular}} & \textbf{\begin{tabular}[c]{@{}c@{}}0.641\\ (-21.4\%)\end{tabular}} \\ \hline
\multirow{4}{*}{OpenDialKG} & iEvaLM                                                      & 0.269                                                     & 0.469                                                     & 0.603                                                     & 0.273                                                     & 0.412                                                     & 0.540                                                     & 0.280                                                     & 0.494                                                     & 0.646                                                     & 0.425                                                     & 0.774                                                     & 0.941                                                     \\
                            & \begin{tabular}[c]{@{}c@{}}iEvaLM\\ (-history)\end{tabular} & \begin{tabular}[c]{@{}c@{}}0.120\\ (-53.3\%)\end{tabular} & \begin{tabular}[c]{@{}c@{}}0.271\\ (-42.2\%)\end{tabular} & \begin{tabular}[c]{@{}c@{}}0.423\\ (-29.8\%)\end{tabular} & \begin{tabular}[c]{@{}c@{}}0.193\\ (-29.3\%)\end{tabular} & \begin{tabular}[c]{@{}c@{}}0.300\\ (-27.2\%)\end{tabular} & \begin{tabular}[c]{@{}c@{}}0.392\\ (-27.4\%)\end{tabular} & \begin{tabular}[c]{@{}c@{}}0.189\\ (-32.5\%)\end{tabular} & \begin{tabular}[c]{@{}c@{}}0.352\\ (-28.7\%)\end{tabular} & \begin{tabular}[c]{@{}c@{}}0.510\\ (-21.1\%)\end{tabular} & \begin{tabular}[c]{@{}c@{}}0.243\\ (-42.8\%)\end{tabular} & \begin{tabular}[c]{@{}c@{}}0.691\\ (-10.7\%)\end{tabular} & \begin{tabular}[c]{@{}c@{}}0.946\\ (+0.5\%)\end{tabular}  \\
                            & \begin{tabular}[c]{@{}c@{}}iEvaLM\\ (-response)\end{tabular}  & \begin{tabular}[c]{@{}c@{}}0.257\\ (-4.5\%)\end{tabular}  & \begin{tabular}[c]{@{}c@{}}0.454\\ (-3.2\%)\end{tabular}  & \begin{tabular}[c]{@{}c@{}}0.580\\ (-3.8\%)\end{tabular}  & \begin{tabular}[c]{@{}c@{}}0.279\\ (+2.2\%)\end{tabular}  & \begin{tabular}[c]{@{}c@{}}0.417\\ (+1.2\%)\end{tabular}  & \begin{tabular}[c]{@{}c@{}}0.526\\ (-2.6\%)\end{tabular}  & \begin{tabular}[c]{@{}c@{}}0.276\\ (-1.4\%)\end{tabular}  & \begin{tabular}[c]{@{}c@{}}0.483\\ (-2.2\%)\end{tabular}  & \begin{tabular}[c]{@{}c@{}}0.616\\ (-4.6\%)\end{tabular}  & \begin{tabular}[c]{@{}c@{}}0.432\\ (+1.6\%)\end{tabular}  & \begin{tabular}[c]{@{}c@{}}0.737\\ (-4.8\%)\end{tabular}  & \begin{tabular}[c]{@{}c@{}}0.922\\ (-2.0\%)\end{tabular}  \\
                            & \textbf{\begin{tabular}[c]{@{}c@{}}iEvaLM\\ (-both)\end{tabular}}    & \textbf{\begin{tabular}[c]{@{}c@{}}0.086\\ (-68.0\%)\end{tabular}} & \textbf{\begin{tabular}[c]{@{}c@{}}0.227\\ (-51.5\%)\end{tabular}} & \textbf{\begin{tabular}[c]{@{}c@{}}0.367\\ (-39.1\%)\end{tabular}} & \textbf{\begin{tabular}[c]{@{}c@{}}0.198\\ (-27.5\%)\end{tabular}} & \textbf{\begin{tabular}[c]{@{}c@{}}0.301\\ (-26.9\%)\end{tabular}} & \textbf{\begin{tabular}[c]{@{}c@{}}0.369\\ (-31.7\%)\end{tabular}} & \textbf{\begin{tabular}[c]{@{}c@{}}0.167\\ (-40.4\%)\end{tabular}} & \textbf{\begin{tabular}[c]{@{}c@{}}0.316\\ (-36.0\%)\end{tabular}} & \textbf{\begin{tabular}[c]{@{}c@{}}0.461\\ (-28.6\%)\end{tabular}} & \textbf{\begin{tabular}[c]{@{}c@{}}0.127\\ (-70.1\%)\end{tabular}} & \textbf{\begin{tabular}[c]{@{}c@{}}0.552\\ (-28.7\%)\end{tabular}} & \textbf{\begin{tabular}[c]{@{}c@{}}0.912\\ (-3.1\%)\end{tabular}}  \\ \hline
\end{tabular}}
\label{table:1}
\end{table*}
\begin{figure*}[htbp]
    \centering
    \includegraphics[width=0.95\textwidth]{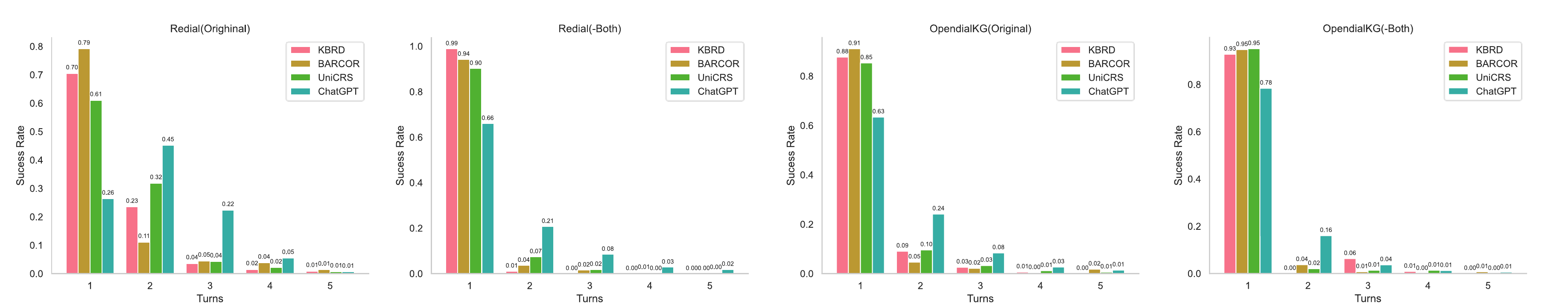}
    \caption{Percentage of successful recommendations by turn when using iEvaLM as the user simulator.}
    \label{fig-4}
\end{figure*}
\subsubsection{RQ1: Does the current user simulator, iEvaLM, exhibit data leakage, and if so, in which process? How does the model perform when we ignore these successful recommendations affected by data leakage?}

\ 
\newline
\indent We first conducted a case study as shown in Figure \ref{fig-2} and Figure \ref{fig-3}, and found that data leakage occurs in the conversational history and user simulator's replies. In Figure \ref{fig-2}, the conversational history, annotated by humans, includes the target item, enabling the CRS to utilize this history to directly achieve a successful recommendation. Conversely, in Figure \ref{fig-3}, the user simulator explicitly discloses the title of the target item during the interaction, thereby allowing the CRS to directly formulate a successful recommendation based on the user’s explicit indication. 

Additionally, we evaluated the performance of different baseline models under various data leakage scenarios. Here, '-history' denotes that in our evaluation, conversations influenced by data leakage from conversational history, resulting in successful recommendations, are excluded from consideration. '-response' denotes that conversations affected by data leakage from the user simulator, which result in successful recommendations, are excluded from consideration. '-both' denotes that scenarios involving data leakage leading to successful recommendations in both aforementioned contexts are not considered in our evaluation. The evaluation results are presented in Table \ref{table:1}. We can get the following observations:
\begin{itemize}
    \item When we exclude these successful recommendation conversations affected by data leakage, it is observed that all the baseline models exhibit a decrease in performance, with the degree of decrease varying among models. For instance, in the case of recall@50, the baseline models show a decrease in the ReDial dataset by 21.6\%, 13.8\%, 13.5\%, and 21.4\% respectively, and a similar decrease by 39.1\%, 31.7\%, 28.6\%, and 3.1\% respectively in the OpenDialKG dataset.
    \item In our experiments, we found that KBRD outperforms BARCOR when exploiting data leakage. However, without considering data leakage, which occurs in the conversational history and user simulator's replies, KBRD ranks the lowest among all baseline models.
    \item The experimental results show that ChatGPT still performs the best, and UniCRS remains the top CRS method for state-of-the-art (SOTA).
\end{itemize}

From the observations, we can conclude that the current simulator, iEvaLM, exhibits data leakage in the conversational history and user simulator's replies, resulting in inflated evaluation results.

\subsubsection{RQ2: How much do successful recommendation conversations depend on user simulator interactions compared to conversational history?}
\ 
\newline
\indent To answer this question, we counted the number of interaction turns used by all successfully recommended conversations. If the CRS successfully recommends on the first turn, it suggests that the CRS can make successful recommendations using only information from conversational history. The evaluation results are presented in Figure \ref{fig-4}, where `Original' denotes that in the evaluation, we consider all conversations with successful recommendations, even if those recommendations may be affected by data leakage in the conversational history and user simulator's replies. `-Both' denotes that in the evaluation, any conversations resulting in successful recommendations that are affected by data leakage, including from conversational history and user simulator's replies, are not considered. There are some observations:
\begin{itemize}
    \item As can be seen in Figure \ref{fig-4}, all models exhibit a very high percentage of successful recommendations in the first round on both datasets, when the CRS has not yet utilized the feedback information given by the user simulator.
    \item  Comparing `Original' and '-Both',  it can be observed that in the '-Both' scenario, the success rate of recommendations by the CRS significantly decreases in rounds 2 to 5. This indicates a challenge for the CRS in effectively utilizing the interaction information provided by the user simulator.
    \item Compared to other baseline models, ChatGPT outperforms them in terms of successful recommendations from rounds 2 through 5. We hypothesize that this is because ChatGPT's extensive world knowledge enables it to utilize conversational information more effectively in generating recommendations.
\end{itemize}

Based on the observations, we can conclude that the success of CRS recommendations depends more on the availability and quality of conversational history than on the output from the user simulator.
\subsubsection{RQ3: Can the user simulator generate responses that meet expectations across various dataset scenarios? If not, why?}
\begin{figure}[htp]
    \includegraphics[width=0.5\textwidth]{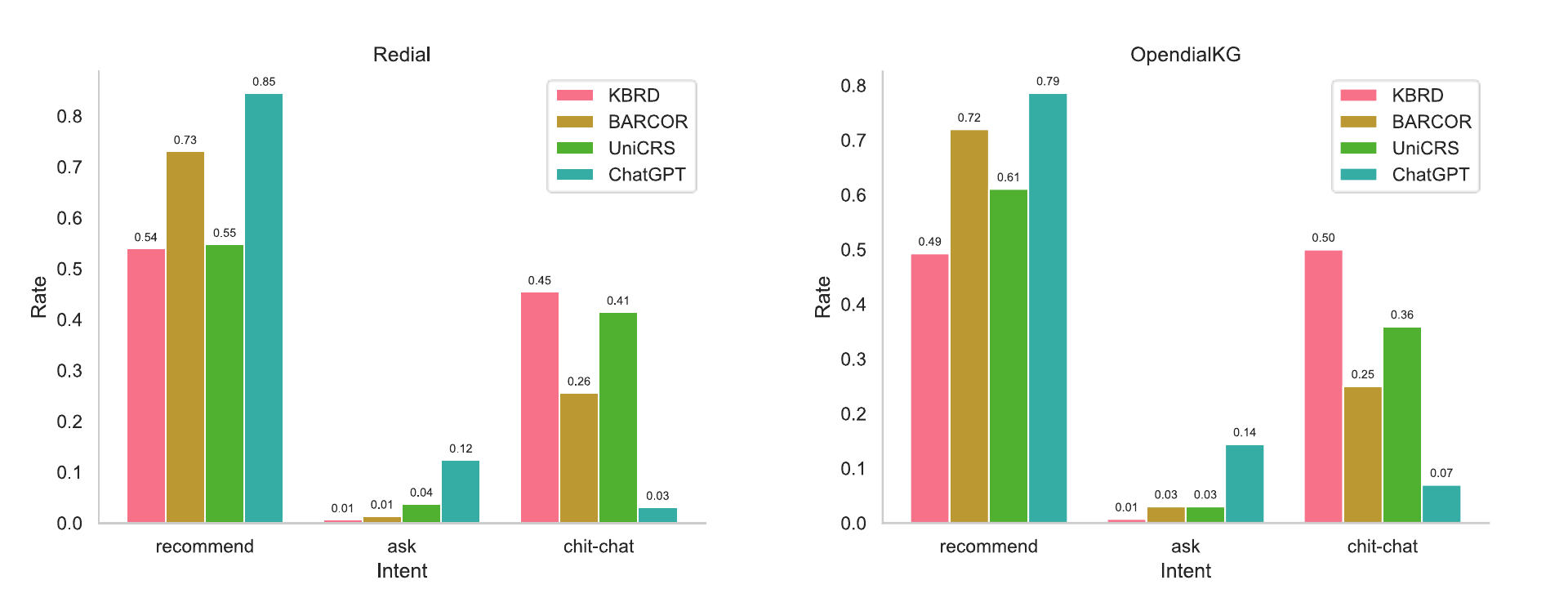}
    \caption{The proportion of the CRS's intents during the interaction.}
    \label{fig-5}
\end{figure}
\ 
\newline
\indent We utilize LLM to understand and quantify the intent distribution of the CRS during interactions, which includes three types of intents: chit-chat, ask, and recommend. `Chit-chat' refers to instances where the CRS did not explicitly ask the user for their preference, an example of this could be a simple greeting like ``Good morning!". `Ask' refers to instances where the CRS explicitly asked the user for their preference, for example, ``Do you have a director in mind?". Lastly, `Recommend' refers to instances where the CRS suggested something to the user, such as recommending a specific movie based on the user's previous preferences. The evaluation results are presented in Figure \ref{fig-5}. Combined with Figure \ref{fig-4}, there are some observations:
\begin{itemize}
    \item From Figure \ref{fig-5}, it can be seen that among the three intents, the proportion of recommend is the largest, while the proportion of ask is very low. Combined with Figure \ref{fig-4}, we speculate that this is because the conversational history contains sufficient information for the CRS to successfully make recommendations without the need to interact with the user simulator, thus achieving success in the first round of recommendations.
    \item During the interaction process, the proportion of chit-chat is relatively high. This is because both the ReDial and OpenDialKG datasets are created in a chit-chat manner. However, current user simulators were not designed with the capability to operate within chit-chat scenarios in mind. As illustrated in Figure \ref{fig-4}, it can be observed that CRS methods do not effectively utilize the interaction information obtained from user simulators.
\end{itemize}

From the observations, we hypothesize that controlling the output of the user simulator through a single prompt template proves challenging, due to the complexity of scenarios and the difficulty in finely controlling the user simulator's response across different contexts through single-prompe. We will conduct preliminary experiments to verify this hypothesis in Chapter 3.

From the experimental phenomena observed, we can identify three significant and non-negligible issues: (1) Data leakage, which occurs in conversational history and the user simulator's replies, results in inflated evaluation results. (2) The success of CRS recommendations depends more on the availability and quality of conversational history than on the responses from user simulators. (3) Controlling the output of the user simulator through a single prompt template proves challenging. These issues underpin our belief that current user simulators are not realistic and trustworthy. In the following section, we will propose a simple user simulator to mitigate the aforementioned problems.

\section{The proposed simple user simulator}
\begin{table*}[htp]
\caption{Performance of CRSs and ChatGPT under SimpleUserSim in various data leakage scenarios.}
\resizebox{\textwidth}{!}{
\begin{tabular}{cc|ccc|ccc|ccc|ccc}
\hline
\multicolumn{2}{c|}{Model}                                                                       & \multicolumn{3}{c|}{KBRD}                                                                                                                                                         & \multicolumn{3}{c|}{BARCOR}                                                                                                                                                       & \multicolumn{3}{c|}{UniCRS}                                                                                                                                                       & \multicolumn{3}{c}{ChatGPT}                                                                                                                                                      \\ \hline
\multicolumn{2}{c|}{datasets}                                                                    & Recall@1                                                  & Recall@10                                                 & Recall@50                                                 & Recall@1                                                  & Recall@10                                                 & Recall@50                                                 & Recall@1                                                  & Recall@10                                                 & Recall@50                                                 & Recall@1                                                  & Recall@10                                                 & Recall@50                                                \\ \hline
\multirow{4}{*}{ReDial}     & SimpleUserSim                                                      & 0.027                                                     & 0.170                                                     & 0.443                                                     & 0.030                                                     & 0.205                                                     & 0.510                                                     & 0.073                                                     & 0.284                                                     & 0.592                                                     & 0.209                                                     & 0.490                                                     & 0.706                                                    \\
                            & \begin{tabular}[c]{@{}c@{}}SimpleUserSim\\ (-history)\end{tabular} & \begin{tabular}[c]{@{}c@{}}0.009\\ (-66.7\%)\end{tabular} & \begin{tabular}[c]{@{}c@{}}0.117\\ (-31.2\%)\end{tabular} & \begin{tabular}[c]{@{}c@{}}0.385\\ (-13.1\%)\end{tabular} & \begin{tabular}[c]{@{}c@{}}0.030\\ (-0.0\%)\end{tabular}  & \begin{tabular}[c]{@{}c@{}}0.195\\ (-4.9\%)\end{tabular}  & \begin{tabular}[c]{@{}c@{}}0.486\\ (-4.7\%)\end{tabular}  & \begin{tabular}[c]{@{}c@{}}0.033\\ (-54.8\%)\end{tabular} & \begin{tabular}[c]{@{}c@{}}0.232\\ (-18.3\%)\end{tabular} & \begin{tabular}[c]{@{}c@{}}0.563\\ (-4.9\%)\end{tabular}  & \begin{tabular}[c]{@{}c@{}}0.179\\ (-14.3\%)\end{tabular} & \begin{tabular}[c]{@{}c@{}}0.464\\ (-5.3\%)\end{tabular}  & \begin{tabular}[c]{@{}c@{}}0.700\\ (-0.8\%)\end{tabular} \\
                            & \begin{tabular}[c]{@{}c@{}}SimpleUserSim\\ (-response)\end{tabular}  & \begin{tabular}[c]{@{}c@{}}0.027\\ (-0.0\%)\end{tabular}  & \begin{tabular}[c]{@{}c@{}}0.170\\ (-0.0\%)\end{tabular}  & \begin{tabular}[c]{@{}c@{}}0.443\\ (-0.0\%)\end{tabular}  & \begin{tabular}[c]{@{}c@{}}0.030\\ (-0.0\%)\end{tabular}  & \begin{tabular}[c]{@{}c@{}}0.205\\ (-0.0\%)\end{tabular}  & \begin{tabular}[c]{@{}c@{}}0.510\\ (-0.0\%)\end{tabular}  & \begin{tabular}[c]{@{}c@{}}0.070\\ (-4.1\%)\end{tabular}  & \begin{tabular}[c]{@{}c@{}}0.276\\ (-2.8\%)\end{tabular}  & \begin{tabular}[c]{@{}c@{}}0.590\\ (-0.3\%)\end{tabular}  & \begin{tabular}[c]{@{}c@{}}0.199\\ (-4.8\%)\end{tabular}  & \begin{tabular}[c]{@{}c@{}}0.484\\ (-1.2\%)\end{tabular}  & \begin{tabular}[c]{@{}c@{}}0.705\\ (-0.1\%)\end{tabular} \\
                            & \textbf{\begin{tabular}[c]{@{}c@{}}SimpleUserSim\\ (-both)\end{tabular}}    & \textbf{\begin{tabular}[c]{@{}c@{}}0.009\\ (-66.7\%)\end{tabular}} & \textbf{\begin{tabular}[c]{@{}c@{}}0.117\\ (-31.2\%)\end{tabular}} & \textbf{\begin{tabular}[c]{@{}c@{}}0.385\\ (-13.1\%)\end{tabular}} & \textbf{\begin{tabular}[c]{@{}c@{}}0.030\\ (-0.0\%)\end{tabular}}  & \textbf{\begin{tabular}[c]{@{}c@{}}0.195\\ (-4.9\%)\end{tabular}}  & \textbf{\begin{tabular}[c]{@{}c@{}}0.486\\ (-4.7\%)\end{tabular}}  & \textbf{\begin{tabular}[c]{@{}c@{}}0.030\\ (-58.9\%)\end{tabular}} & \textbf{\begin{tabular}[c]{@{}c@{}}0.223\\ (-21.4\%)\end{tabular}} & \textbf{\begin{tabular}[c]{@{}c@{}}0.560\\ (-5.4\%)\end{tabular}}  & \textbf{\begin{tabular}[c]{@{}c@{}}0.169\\ (-19.1\%)\end{tabular}} & \textbf{\begin{tabular}[c]{@{}c@{}}0.458\\ (-6.5\%)\end{tabular}}  & \textbf{\begin{tabular}[c]{@{}c@{}}0.700\\ (-0.8\%)\end{tabular}} \\ \hline
\multirow{4}{*}{OpenDialKG} & SimpleUserSim                                                      & 0.243                                                     & 0.432                                                     & 0.558                                                     & 0.276                                                     & 0.423                                                     & 0.545                                                     & 0.256                                                     & 0.458                                                     & 0.614                                                     & 0.429                                                     & 0.724                                                     & 0.918                                                    \\
                            & \begin{tabular}[c]{@{}c@{}}SimpleUserSim\\ (-history)\end{tabular} & \begin{tabular}[c]{@{}c@{}}0.079\\ (-67.5\%)\end{tabular} & \begin{tabular}[c]{@{}c@{}}0.213\\ (-50.7\%)\end{tabular} & \begin{tabular}[c]{@{}c@{}}0.353\\ (-36.7\%)\end{tabular} & \begin{tabular}[c]{@{}c@{}}0.201\\ (-27.2\%)\end{tabular} & \begin{tabular}[c]{@{}c@{}}0.315\\ (-25.5\%)\end{tabular} & \begin{tabular}[c]{@{}c@{}}0.397\\ (-27.2\%)\end{tabular} & \begin{tabular}[c]{@{}c@{}}0.152\\ (-40.6\%)\end{tabular} & \begin{tabular}[c]{@{}c@{}}0.296\\ (-35.3\%)\end{tabular} & \begin{tabular}[c]{@{}c@{}}0.472\\ (-23.1\%)\end{tabular} & \begin{tabular}[c]{@{}c@{}}0.255\\ (-40.6\%)\end{tabular} & \begin{tabular}[c]{@{}c@{}}0.604\\ (-16.6\%)\end{tabular} & \begin{tabular}[c]{@{}c@{}}0.909\\ (-1.0\%)\end{tabular} \\
                            & \begin{tabular}[c]{@{}c@{}}SimpleUserSim\\ (-response)\end{tabular}  & \begin{tabular}[c]{@{}c@{}}0.243\\ (-0.0\%)\end{tabular}  & \begin{tabular}[c]{@{}c@{}}0.432\\ (-0.0\%)\end{tabular}  & \begin{tabular}[c]{@{}c@{}}0.558\\ (-0.0\%)\end{tabular}  & \begin{tabular}[c]{@{}c@{}}0.281\\ (+1.8\%)\end{tabular}  & \begin{tabular}[c]{@{}c@{}}0.426\\ (+0.7\%)\end{tabular}  & \begin{tabular}[c]{@{}c@{}}0.540\\ (-0.9\%)\end{tabular}  & \begin{tabular}[c]{@{}c@{}}0.260\\ (+1.6\%)\end{tabular}  & \begin{tabular}[c]{@{}c@{}}0.463\\ (+1.1\%)\end{tabular}  & \begin{tabular}[c]{@{}c@{}}0.608\\ (-1.0\%)\end{tabular}  & \begin{tabular}[c]{@{}c@{}}0.391\\ (-8.9\%)\end{tabular}  & \begin{tabular}[c]{@{}c@{}}0.680\\ (-6.1\%)\end{tabular}  & \begin{tabular}[c]{@{}c@{}}0.904\\ (-1.5\%)\end{tabular} \\
                            & \textbf{\begin{tabular}[c]{@{}c@{}}SimpleUserSim\\ (-both)\end{tabular}}    & \textbf{\begin{tabular}[c]{@{}c@{}}0.079\\ (-67.5\%)\end{tabular}} & \textbf{\begin{tabular}[c]{@{}c@{}}0.213\\ (-50.7\%)\end{tabular}} & \textbf{\begin{tabular}[c]{@{}c@{}}0.353\\ (-36.7\%)\end{tabular}} & \textbf{\begin{tabular}[c]{@{}c@{}}0.203\\ (-26.4\%)\end{tabular}} & \textbf{\begin{tabular}[c]{@{}c@{}}0.317\\ (-25.1\%)\end{tabular}} & \textbf{\begin{tabular}[c]{@{}c@{}}0.392\\ (-28.1\%)\end{tabular}} & \textbf{\begin{tabular}[c]{@{}c@{}}0.154\\ (-39.8\%)\end{tabular}} & \textbf{\begin{tabular}[c]{@{}c@{}}0.301\\ (-34.3\%)\end{tabular}} & \textbf{\begin{tabular}[c]{@{}c@{}}0.463\\ (-24.6\%)\end{tabular}} & \textbf{\begin{tabular}[c]{@{}c@{}}0.138\\ (-67.8\%)\end{tabular}} & \textbf{\begin{tabular}[c]{@{}c@{}}0.496\\ (-31.5\%)\end{tabular}} & \textbf{\begin{tabular}[c]{@{}c@{}}0.882\\ (-3.9\%)\end{tabular}} \\ \hline
\end{tabular}}
\label{table:2}
\end{table*}
\begin{figure*}[htbp]
    \centering
    \includegraphics[width=0.95\textwidth]{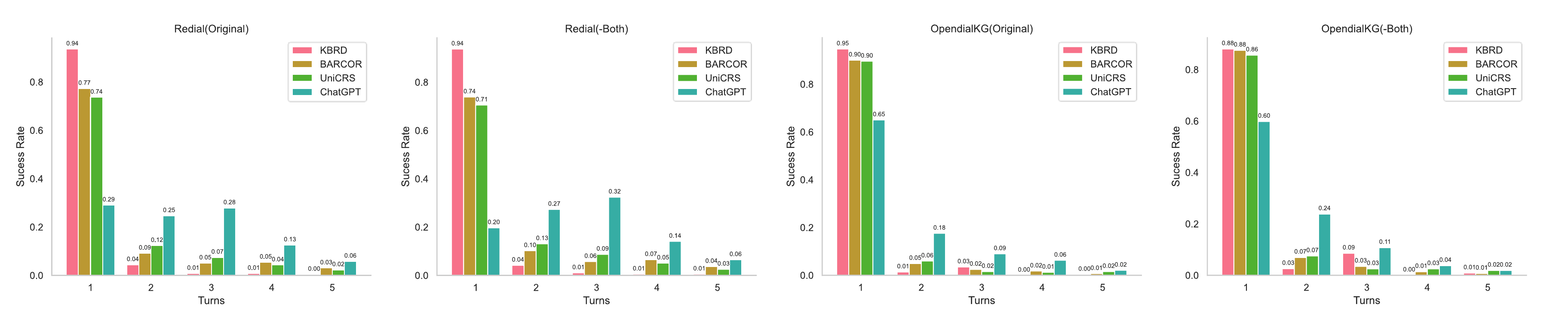}
    \caption{Percentage of successful recommendation by turn when using SimpleUserSim for user simulation.}
    \label{fig-6}
\end{figure*}
\subsection{A very intuitive improvement}
Based on the observations, we believe that a more realistic and trustworthy user simulator should emulate human cognitive processes. Therefore, in this section, we propose a very intuitive improvement, which is named SimpleUserSim. To mitigate the issues mentioned above, SimpleUserSim mainly introduces two improvements compared to iEvaLM: (1) SimpleUserSim ensures that, during the conversation, the simulator is only aware of the target items' attribute information. This means that, until a successful recommendation is made, the simulator does not know the target items' titles. (2) Based on the CRS's intent, the following three actions are taken:
\begin{itemize}
    \item Chit-chat: SimpleUserSim generates a flow of responses based on the current topic combined with current preferences.
    \item Ask: SimpleUserSim responds to the CRS's questions based on real-time preferences.
    \item Recommend: SimpleUserSim checks if the recommended items align with its target items and provides positive or negative feedback accordingly.
\end{itemize}
It's worth mentioning that we use different prompts to control various actions of the user simulator.

\subsection{Experiment}
In this experiment, we replaced the existing user simulator with our SimpleUserSim, while maintaining the same experimental setup. The evaluation results, depicted in Table \ref{table:2} and Figure \ref{fig-6}, yield the following observations:
\begin{itemize}
    \item Data leakage from conversational history is an inherent problem of the dataset and cannot be solely addressed through the user simulator.
    \item As observed in Table \ref{table:2}, our approach significantly mitigates the data leakage issue caused by the user simulator. For instance, when using iEvaLM as the user simulator, different CRS methods experienced a decrease in the recall@50 metric on the ReDial dataset by 12.0\%, 6.8\%, 10.3\%, and 17.9\%, respectively. In contrast, when using our user simulator under the same experimental setup, the metrics decreased by 0\%, 0\%, 0.3\%, and 0.1\%, respectively.
    \item Despite the user simulator not knowing the target items' titles, it can still inadvertently leak those titles. This leakage occurs because the user simulator enriches the conversation with information from its own world knowledge when expressing preferences. Inadvertently, this may include the target item titles, for example, when listing a favorite director's masterpieces to the CRS.
    \item As illustrated in Figure \ref{fig-6}, in the `-Both' scenario, our approach demonstrates superior performance across multiple rounds of interactions (from the 2nd to the 5th round). This superior performance is attributable to SimpleUserSim's ability to express its preferences in chit-chat conversational scenarios, thereby enabling the CRS to more effectively utilize the user simulator's responses for recommendations.
\end{itemize}

\section{CONCLUSION AND DISCUSSION}
In this work, we analyze specific limitations associated with the use of LLMs in constructing user simulators for CRS. In addition to this analysis, we propose a solution, SimpleUserSim, which effectively addresses the issue of user simulators inadvertently revealing the target item titles. SimpleUserSim guides CRSs to better utilize conversational information to make successful recommendations. We hope that our experimental findings can provide valuable insights for future research in this field.

Even without fine-tuning, LLM exhibits excellent performance in terms of accuracy and speed. Given its superior content/context knowledge, LLM holds great promise as an effective approach for CRS tasks. At the same time, it is also worth exploring how LLM can be utilized to build realistic and trustworthy user simulators. These simulators can provide a more comprehensive evaluation of CRS, including its performance under different scenarios and its ability to handle complex tasks.

\section*{Acknowledgments}
This work was supported in part by the National Key Research and Development Program of China under Grant (2023YFC3310700), the National Natural Science Foundation of China (62202041), the Fundamental Research Funds for the Central Universities (2023JBM
C057).
\bibliographystyle{unsrt}
\bibliography{sample-base}

\end{document}